\documentclass[sigconf,natbib=true]
{acmart}
\usepackage{bbm}
\usepackage{amsmath}
\usepackage{array}
\usepackage{multirow}
\AtBeginDocument{%
  }

\setcopyright{acmlicensed}
\copyrightyear{2018}
\acmYear{2018}
\acmDOI{XXXXXXX.XXXXXXX}
\acmConference[Conference acronym 'XX]{Make sure to enter the correct
  conference title from your rights confirmation email}{June 03--05,
  2018}{Woodstock, NY}
\acmISBN{978-1-4503-XXXX-X/2018/06}

\begin{document}
\title{Graph-Based Multimodal Contrastive Learning for Chart Question Answering}

\author{Yue Dai}
\email{yue.dai@research.uwa.edu.au}
\orcid{0000-0003-1160-7927}
\affiliation{%
  \institution{The University of Western Australia}
  \streetaddress{P.O. Box 1212}
  \city{Perth}
  \state{Western Australia}
  \country{Australia}
  \postcode{43017-6221}
}

\author{Soyeon Caren Han}
\authornote{Corresponding author.}
\affiliation{%
  \institution{The University of Melbourne}
  \streetaddress{1 Th{\o}rv{\"a}ld Circle}
  \city{Melbourne}
  \country{Australia}}
\email{caren.han@unimelb.edu.au}

\author{Wei Liu}
\affiliation{%
  \institution{The University of Western Australia}
  \streetaddress{1 Th{\o}rv{\"a}ld Circle}
  \city{Perth}
  \country{Australia}}
\email{wei.liu@uwa.edu.au}

\renewcommand{\shortauthors}{Trovato et al.}

\begin{abstract}
% Chart Question Answering (ChartQA) presents significant challenges due to the complex distribution of chart elements and the implicit patterns embedded within the underlying data. In this chapter, we have developed a joint multimodal scene graph for charts, explicitly representing the relationships between chart elements and their associated patterns.
% Our proposed multimodal scene graph consists of two components: a visual graph and a textual graph, each designed to capture the structural and semantic information within the chart. To unify representations across these different modalities, we introduce a multimodal graph contrastive learning approach that learns unified representations by maximizing similarity between nodes representing the same object across multimodal graphs. The learned graph representations can be seamlessly incorporated into a transformer decoder as a soft prompt.
% Additionally, given the growing need for Multimodal Large Language Models (MLLMs) in zero-shot scenarios, we have designed Chain-of-Thought (CoT) prompts for MLLMs to reduce hallucinations. We tested both methods on public benchmarks such as ChartQA, OpenCQA, and ChartX, demonstrating improved performance and validating the effectiveness of our proposed methods.
Chart question answering (ChartQA) is challenged by the heterogeneous composition of chart elements and the subtle data patterns they encode. This work introduces a novel joint multimodal scene graph framework that explicitly models the relationships among chart components and their underlying structures. The framework integrates both visual and textual graphs to capture structural and semantic characteristics, while a graph contrastive learning strategy aligns node representations across modalities—enabling their seamless incorporation into a transformer decoder as soft prompts. Moreover, a set of tailored Chain-of-Thought (CoT) prompts is proposed to enhance multimodal large language models (MLLMs) in zero-shot scenarios by mitigating hallucinations. Extensive evaluations on benchmarks including ChartQA, OpenCQA, and ChartX demonstrate significant performance improvements and validate the efficacy of the proposed approach.
\end{abstract}



\keywords{Do, Not, Us, This, Code, Put, the, Correct, Terms, for,
  Your, Paper}

\maketitle

\section{Introduction}
In the era of big data, a huge amount of data is generated and utilized daily. Charts are among the most effective tools for data analysis, but understanding charts is a non-trivial task. Significant effort is required to extract values and perform numerical reasoning. Automating chart understanding could substantially enhance the productivity of professionals who work with data regularly. One critical task in chart understanding is ChartQA, where an agent must answer questions based on its comprehension of a chart. Significant research has been devoted to this area. Early works \cite{DBLP:conf/acl/MasryLTJH22, DBLP:conf/acl/0001PKPLJACE23, DBLP:conf/emnlp/MasryKLHJ23} relied on fine-tuning models on specific datasets. With the rise of instruction tuning, numerous Vision-and-Language Pre-trained Models (VLPMs) \cite{DBLP:conf/acl/0001PKPLJACE23, DBLP:conf/emnlp/MasryKLHJ23, han2023chartllama, meng-etal-2024-chartassistant} and Multimodal Large Language Models (MLLMs) \cite{wang2024qwen2, DBLP:journals/corr/abs-2407-21783} have demonstrated the ability to perform chart understanding tasks in a zero-shot setting.
Most VLPMs and MLLMs utilize transformer-based architectures, which segment an image input into patches, thus losing object-level information. 
Previous works \cite{DBLP:conf/aaai/LiuWL24, DBLP:conf/cvpr/GaoL0SC20, DBLP:conf/aaai/WangYLZPLCC22} addressed this challenge in natural image understanding by employing multimodal scene graphs to represent objects and their relationships. Inspired by this, we propose leveraging multimodal scene graphs for chart understanding. A key challenge in multimodal representation learning is the effective fusion of representations from different modalities. While approaches like simple concatenation, max pooling, and linear projection have been used, contrastive learning has gained prominence following the success of CLIP \cite{DBLP:conf/icml/RadfordKHRGASAM21}. Although contrastive learning has been applied in graph representation learning \cite{DBLP:conf/iclr/VelickovicFHLBH19, DBLP:conf/nips/0001XLW21}, these methods are typically limited to single-modality settings.
In this work, we introduce a novel multimodal graph contrastive learning method. The learned unified graph representation is integrated as a soft prompt for the decoder, concatenated before the text input. Beyond fine-tuning pre-trained models, we explore zero-shot capabilities on MLLMs. Instruction tuning enables Large Language Models (LLMs) and MLLMs to tackle diverse tasks without requiring fine-tuning. However, a significant limitation of LLMs is hallucination \cite{DBLP:journals/csur/JiLFYSXIBMF23}. Inspired by \cite{DBLP:conf/nips/Wei0SBIXCLZ22}, we investigate the potential of chain-of-thought (CoT) prompting for VLPMs and MLLMs in ChartQA. 
Our contributions in this chapter can be summarized as follows. First, we propose a novel graph contrastive learning method tailored for multimodal scene graphs in chart understanding. This approach incorporates a multimodal scene graph layer that integrates seamlessly with transformer-based models as a learnable soft prompt. Second, we demonstrate through comprehensive experiments that the proposed graph layer, when combined with graph contrastive learning, significantly boosts the performance of backbone models on widely used datasets, including ChartQA and OpenCQA. Finally, we investigate the application of chain-of-thought (CoT) prompting in vision-language pre-trained models (VLPMs) and multimodal large language models (MLLMs) for chart question answering. Our results reveal the limitations of MLLMs on ChartQA, further validating the necessity of our proposed graph-based method.
\section{Methodology}
\begin{figure}
    \centering
    \includegraphics[width=0.65\linewidth]{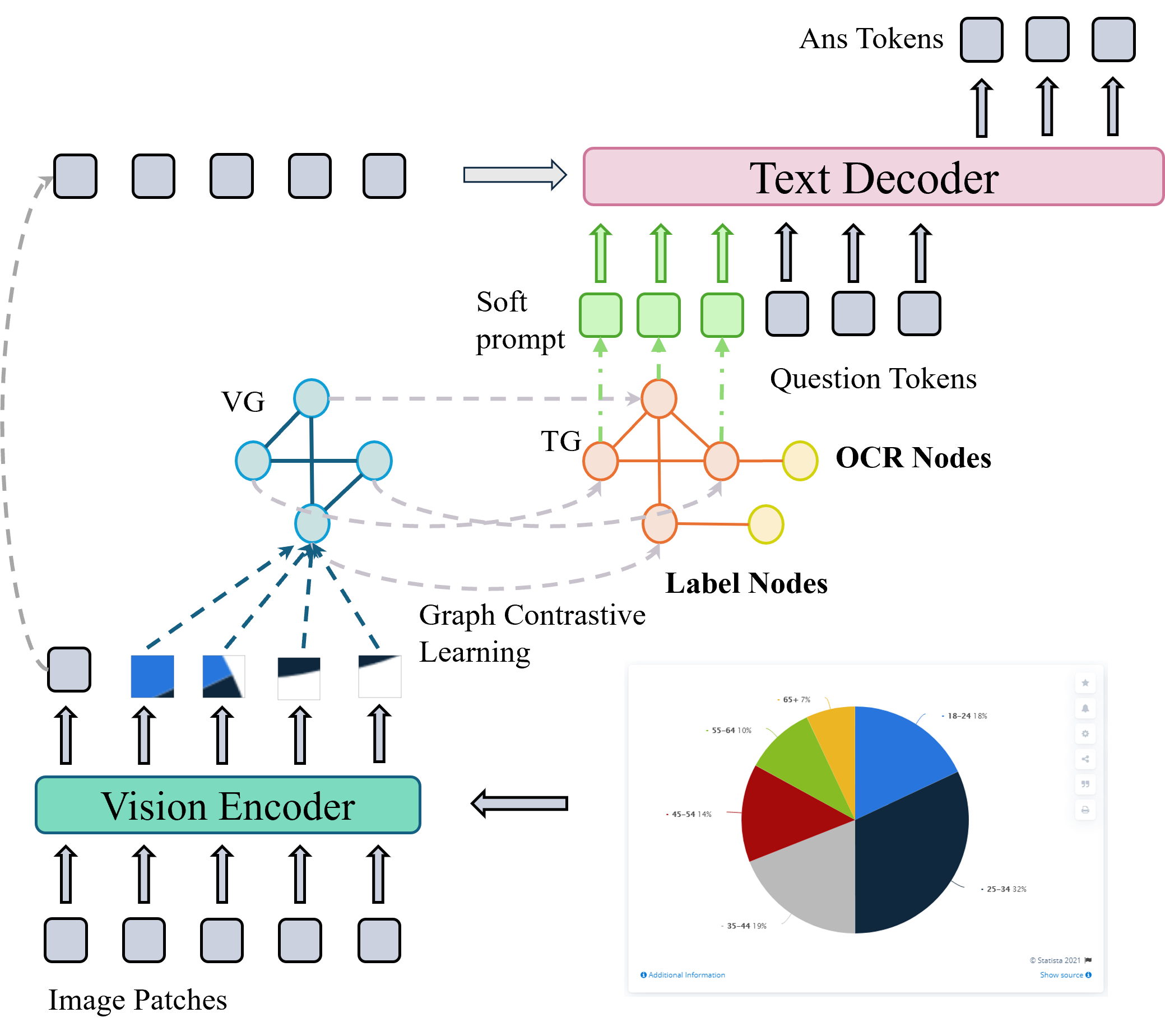}
    \caption{Multimodal Graph Contrastive Learning Prompt with Vision Encoder-Text Decoder (VG: Visual Graph, TG: Textual Graph, Backbone:
UniChart~\cite{DBLP:conf/emnlp/MasryKLHJ23}))}
    \label{fig:unichart-arch}
\end{figure}
\begin{figure}
    \centering
    \includegraphics[width=0.65\linewidth]{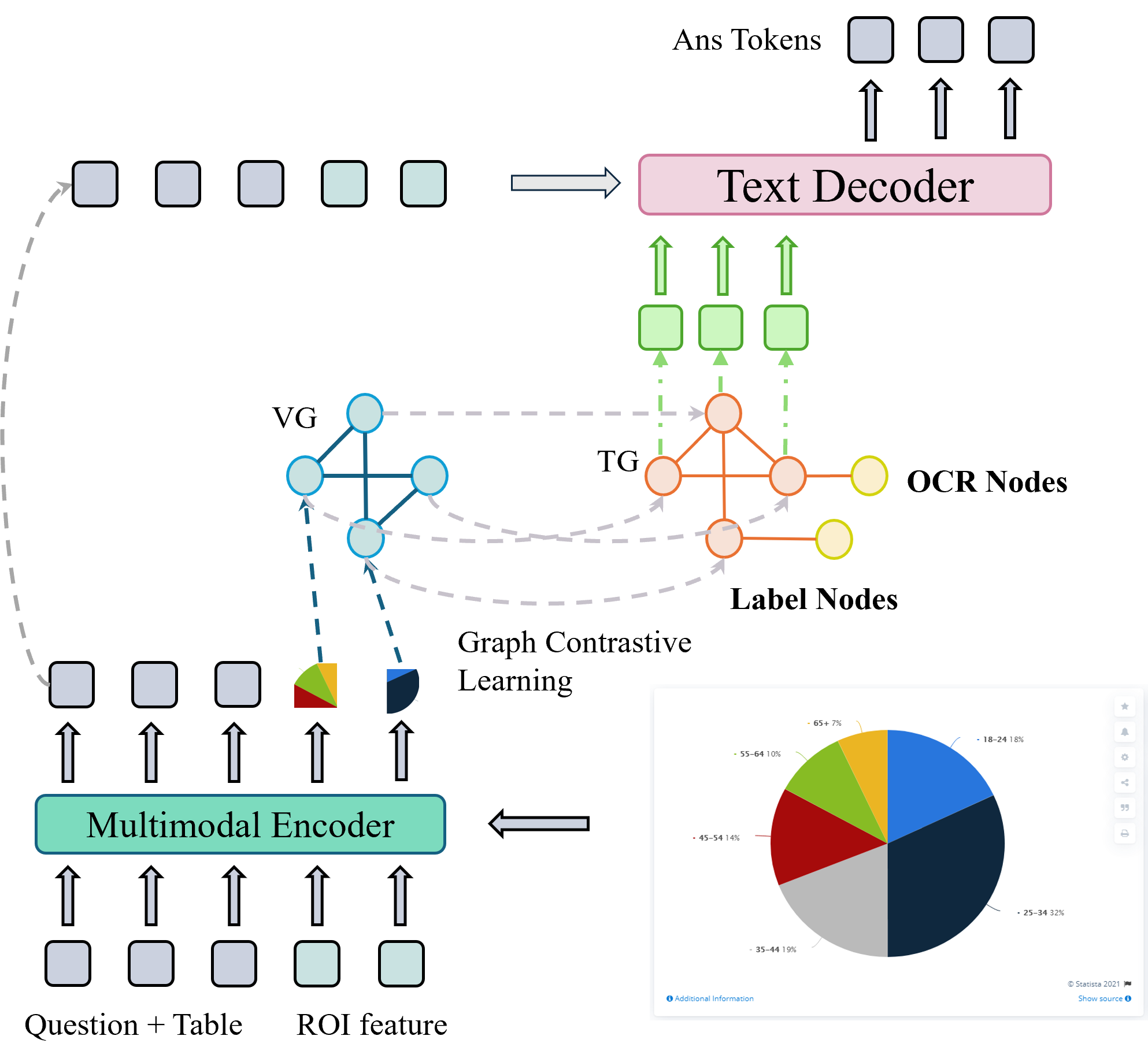}
    \caption{Multimodal Graph Contrastive Learning Prompt with Multimodal Encoder-Text Decoder (VG: Visual Graph, TG: Textual Graph, Backbone:
VL-T5~\cite{DBLP:conf/icml/ChoLTB21})}
    \label{fig:vlt5-arch}
\end{figure}

% \begin{figure}[!h]
%     \centering
%     \begin{subfigure}[b]{0.7\linewidth}
%         \centering
%         \includegraphics[width=\linewidth]{Chapter_4/imgs/unichart_arch.png}
%         \caption{Vision Encoder-Text Decoder (Backbone:
% UniChart~\cite{DBLP:conf/emnlp/MasryKLHJ23})}
%         \label{fig:unichart-arch}
%     \end{subfigure}%
%     \\
%     \hspace{1cm}
%     \centering
%     \begin{subfigure}[b]{0.7\linewidth}
%         \includegraphics[width=\linewidth]{Chapter_4/imgs/vlt5_arch.png}
%         \caption{Multimodal Encoder-Text Decoder
% (Backbone: VL-T5~\cite{DBLP:conf/icml/ChoLTB21})}
%         \label{fig:vlt5-arch}
%     \end{subfigure}
%     \caption{Multimodal Graph Contrastive Learning Prompt Architecture (VG: Visual Graph, TG: Textual Graph)}
%     \label{fig:msggcl_arch}
% \end{figure}

Figure \ref{fig:unichart-arch} and \ref{fig:vlt5-arch} illustrate the proposed graph-based model on UniCha-rt \cite{DBLP:conf/emnlp/MasryKLHJ23} and VL-T5~\cite{DBLP:conf/icml/ChoLTB21}. The models learn graph representations via proposed multimodal graph contrastive learning. The unified multimodal representations are then integrated into the decoder as a soft prompt. This section explains our proposed architecture in detail.

\subsection{Pre-processing}
To initialize the graph where nodes represent objects within charts, the chart images must first undergo preprocessing for object detection. In this work, we utilize Mask R-CNN~\cite{DBLP:conf/iccv/HeGDG17} fine-tuned by~\cite{DBLP:conf/acl/MasryLTJH22} for such task. Once detected, the objects are arranged in a top-down, left-right order. Using the bounding boxes of the detected objects and the predefined patch size of UniChart's final layer (32 $\times$ 32), we compute the alignment between objects and patch indices, which is later used to initialize the visual features of the visual graph. The visual features of each object extracted from Mask R-CNN are used as input for VL-T5. We then apply the state-of-the-art Google OCR model\footnote{\url{https://cloud.google.com/vision/docs/ocr}} to extract text from the chart and align it with the detected objects. Textual features from labels and OCR text are extracted using the CLS token from BERT~\cite{DBLP:conf/naacl/DevlinCLT19}, and these features are used to construct the textual graph.

\subsection{Graph Construction}
\label{sec:gc}
The graphs proposed in this work are inspired by \cite{dai2024msg}, but with some key differences. Below, we outline the graph construction methods used in this paper.

\textbf{Visual Graph} ($G_v$) aims to capture spatial relationships between objects from a visual perspective. Unlike the fully connected graph in \cite{dai2024msg}, we employ a K-nearest neighbors (KNN) approach to highlight the spatial proximity between objects better. Specifically, each node is linked to its three closest neighbors, ensuring that messages are only exchanged between spatially adjacent nodes. Edge weights are calculated as $a_e = \exp(-d)$, where $d$ is the minimum Euclidean distance between the bounding boxes of two objects. For UniChart, nodes are initialized as $V_o = \frac{1}{|P_o|} \sum_{p \in P_o}^{P_o} H_{e_p}$, where $P_o$ represents the set of patches containing object $o$ (obtained during preprocessing), and $H_e$ refers to the final hidden states from the backbone model's encoder. While for VL-T5, since the visual inputs of the encoder are the ROI feature of each object, we directly use the corresponding last hidden states to initialize the nodes in visual graph.

\textbf{Textual Graph} ($G_t$) is constructed using textual features derived from chart elements. Its structure follows that of \cite{dai2024msg}. The node initialization for the textual graph is based on label and OCR features extracted during preprocessing.

\subsection{Multimodal Graph Contrastive Learning}
\label{sec:mgcl}
Contrastive learning has recently gained popularity as a method for learning unified representations across different modalities. Inspired by the multimodal contrastive learning method in \cite{DBLP:conf/icml/RadfordKHRGASAM21}, we aim to maximize the similarity between node pairs from the visual and textual graphs. Specifically, for graphs $G_v$ and $G_t$, using GCN encoders for the visual and textual graphs, denoted as $GCN_v$ and $GCN_t$, we extract the original graph representations $H_v = GCN_v(G_v)$ and $H_t = GCN_t(G_t)$. Following \cite{DBLP:conf/www/0001XYLWW21}, we employ InfoNCE loss as a contrastive objective. Specifically, given the InfoNCE loss equation:

\begin{equation} 
\ell\left(\boldsymbol{u}_i, \boldsymbol{v}_i\right)=\log \frac{e^{\theta\left(\boldsymbol{u}_i, \boldsymbol{v}_i\right) / \tau}}{e^{\theta\left(\boldsymbol{u}_i, \boldsymbol{v}_i\right) / \tau} + \sum{k=1}^N \mathbbm{1}_{[k \neq i]} e^{\theta\left(\boldsymbol{u}_i, \boldsymbol{v}_k\right) / \tau}}
\end{equation}

where $\mathbbm{1}$ is the indicator function that equals 1 if $k \neq i$. The node pairs $u_i$ and $v_i$ are the nodes representing the same object from the visual and textual graphs, excluding the OCR node. For instance, if $u_i$ is the updated node representation based on a bar's visual feature in the visual graph, $v_i$ corresponds to the updated node representation based on the textual feature of "bar" in the textual graph. Consequently, the multimodal graph contrastive learning loss is given by:

\begin{equation}
    \begin{aligned}
    \ell_{cl}  &= \ell_{vt}(u_i, v_i| u \in H_v, v \in \widetilde{H_t})
    \end{aligned}
    \label{eq:loss}
\end{equation}

\subsection{Graph Integration and ChartQA}
Unlike \cite{dai2024msg}, where the multimodal graph representation is integrated into the encoder, we inject the unified multimodal graph representations into the decoder as a soft prompt which has shown promising empirical improvements of helping LLM understanding vision inputs. Specifically, given a pre-trained model consisting of a Vision Encoder and Text Decoder, we learn the visual and textual graph representation as described in Section \ref{sec:gc} and \ref{sec:mgcl}. Then the textual graph representation is incorporated as a soft prompt within the decoder. Specifically, we prepend 36 special `\textlangle G\textrangle' tokens to the question text before passing it into the decoder, and then replace the embeddings of these `\textlangle G\textrangle' tokens with the object node representations from the textual graph. 

\subsection{Chain of Thought Prompting}
We further explored the potential of using CoT (Chain-of-Thought) prompting for MLLMs. Rather than relying on external models to generate supplementary information, such as data tables as shown in~\cite{DBLP:conf/acl/0001EPKPLJCCA23,DBLP:journals/corr/abs-2402-12185}, we prompt the model itself to reason step-by-step, adopting the methodology proposed in~\cite{DBLP:conf/nips/Wei0SBIXCLZ22}. This approach ensures that the self-generated supplementary information is in a format directly understandable by the model. The prompts employed are detailed in Table~\ref{tab:cqaprompt}. Inspired by~\cite{DBLP:conf/acl/0001EPKPLJCCA23,DBLP:journals/corr/abs-2402-12185}, the first prompt instructs the model to convert the chart into a data table before answering the question. Motivated by~\cite{DBLP:conf/acl/TangBS23}, the second prompt treats the chart as a scene graph for reasoning. The third prompt guides the model in decomposing a question into multiple subquestions for step-by-step resolution.

The prompts instruct the model to generate reasoning sentences in the output, which poses challenges in testing the accuracy of the answers. To mitigate this, we introduce a subsequent prompt using the output from the initial reasoning step. The instruction is: ``Reasoning Steps: \textlangle Output\textrangle ~Based on the chart and reasoning step, generate the answer directly.'' We use the chart as the visual input for both the initial and subsequent steps of prompting.

\begin{table}[h]
\centering
\small
\begin{tabular}{p{0.2\linewidth}|p{0.75\linewidth}}
\toprule
\textbf{Prompt No.} & \textbf{Prompt Detail} \\ \midrule
Prompt 1 & Q + Let's first convert the chart to table, and then think step by step \\ \midrule
Prompt 2 & Q + Let's first convert the chart to scene graph, and then think step by step \\ \midrule
Prompt 3 & What are the steps required to answer the following question? + Q \\ \bottomrule
\end{tabular}
\caption{Structured Prompts for Chain of Thought in Chart Question Answering}
\label{tab:cqaprompt}
\vspace{-0.5cm}
\end{table}

\section{Experiment}
\subsection{Dataset and Evaluation Metrics}
We evaluated our GCL model using two datasets: ChartQA and OpenCQA. For CoT prompting, we test the MLLMs on ChartQA and ChartX~\cite{DBLP:journals/corr/abs-2402-12185}. The reason we didn't use OpenCQA is that it's hard for MLLMs to generate similar long answers presented in the dataset with zero-shot of few-shot techniques, which will make BLEU score very low and meaningless. 

\noindent\textbf{ChartQA} is recognized as one of the most demanding datasets for Chart Question Answering. The dataset contains questions generated by either a fine-tuned T5 model ~\cite{DBLP:journals/jmlr/RaffelSRLNMZLL20} or human annotations, leading to two distinct subsets: the augmentation set and the human set. The human set poses greater difficulty, focusing more on tasks involving logical inference and visual reasoning. In line with prior works ~\cite{DBLP:conf/acl/0001PKPLJACE23,DBLP:conf/acl/MasryLTJH22,DBLP:conf/wacv/MethaniGKK20}, we adopt the relaxed accuracy metric, which requires exact matches for text-based answers and applies a tolerance of 5\% for numerical responses.

\noindent\textbf{OpenCQA}~\cite{DBLP:conf/emnlp/KantharajDLTHJ22} is an open-ended dataset within the Chartqa domain. Unlike other Chartqa datasets, where answers typically consist of words or short phrases, responses in OpenCQA are more comprehensive, taking the form of explanatory texts with an average length of 56 tokens. For evaluation, we utilize BLEU4 ~\cite{DBLP:conf/wmt/Post18}, consistent with the approach used in ~\cite{DBLP:conf/emnlp/KantharajDLTHJ22,DBLP:conf/emnlp/MasryKLHJ23}, scaling the score from 0 to 100 to align with previous research.

\noindent\textbf{ChartX} is a recent benchmark designed for chart understanding, comprising 7 chart-related tasks and 18 chart types. In this study, we focus exclusively on the QA task provided. The original paper employed the GPT-acc metric for QA tasks, allowing a 5\% tolerance for numerical answers, similar to relaxed accuracy. However, as our goal of employing CoT prompting is to mitigate hallucination, we argue that using exact match metrics for numerical and textual answers is more appropriate, as demonstrated in~\cite{DBLP:conf/cvpr/LiJTG24}.

\subsection{Overall Performance}
\begin{table}[h]
\centering
\small
\begin{tabular}{p{0.19\linewidth}>{\centering\arraybackslash}p{0.04\linewidth}>{\centering\arraybackslash}p{0.04\linewidth}>{\centering\arraybackslash}p{0.04\linewidth}>{\centering\arraybackslash}p{0.08\linewidth}>{\centering\arraybackslash}p{0.08\linewidth}>{\centering\arraybackslash}p{0.08\linewidth}>{\centering\arraybackslash}p{0.12\linewidth}}
\toprule
\multirow{2}{*}{\textbf{Models}} & \multirow{2}{*}{\textbf{GT}} & \multirow{2}{*}{\textbf{G.}} & \multirow{2}{*}{\textbf{GCL}}  & \multicolumn{3}{c}{\textbf{ChartQA}} & \textbf{OpenCQA} \\
 &  &  &  & \textbf{aug.} & \textbf{human} & \textbf{avg.} & \textbf{BLEU} \\
\midrule
Pix2Struct & $\times$ & $\times$ & $\times$ & 81.60 & 30.50 & 56.00 & - \\
Matcha\footnotemark[3] & $\times$ & $\times$ & $\times$ & 90.20 & 38.20 & 64.20 & - \\
Matcha\footnotemark[4] & $\times$ & $\times$ & $\times$ & 86.64 & 36.96 & 61.80 & - \\ 
\midrule
Qwen2VL Ori & $\times$ & $\times$ & $\times$ & \textbf{85.60} & - & - & 0.64 \\
Qwen2VL P1 & $\times$ & $\times$ & $\times$ & 84.56 & - & - & 0.77 \\
\midrule
UniChart\footnotemark[3] & $\times$ & $\times$ & $\times$ & \underline{88.56} & \underline{43.92} & \underline{66.24} & 14.88 \\
UniChart\footnotemark[4] & $\times$ & $\times$ & $\times$ & 82.32 & 34.48 & 58.40 & 8.76 \\
UniChart\cite{dai2024msg} & $\times$ & $\checkmark$ & $\times$ & 85.36 & 37.44 & 61.4 & 11.91 \\
\textbf{UniChart (O)} & $\times$ & $\checkmark$ & $\checkmark$ & \textbf{90.00} & \textbf{44.88} & \textbf{67.44} & \textbf{16.35} \\ \midrule
VL-T5 & $\checkmark$ & $\times$ & $\times$ & - & - & 59.12 & 14.73 \\ 
VL-T5\cite{dai2024msg} & $\checkmark$ & $\checkmark$ & $\times$ & \textbf{92.4} & \textbf{38} & \textbf{64.8} & 17.14 \\
\textbf{VL-T5 (O)} & $\checkmark$ & $\checkmark$ & $\checkmark$ & \underline{92.00} & \underline{34.96} & \underline{63.60} & \textbf{18.42} \\
\bottomrule
\end{tabular}
\caption{Comprehensive Evaluation of Model Performance on ChartQA and OpenCQA (GT: Ground Truth Table; GCL: multimodal Graph Contrastive Learning; G.: Graph; (O): Ours; Ori: Original QA; P1: Prompt 1 (prompt with best performance))}
\label{tab:gcl_all}
\vspace{-0.5cm}
\end{table}
\footnotetext[3]{Result from original paper}
\footnotetext[4]{Result from checkpoint in huggingface}
The overall performance of our models with the proposed Graph Contrastive Learning (GCL) method is presented in Table \ref{tab:gcl_all}. By comparing results with the backbone models, we observe that, in most cases, the models with GCL demonstrate superior performance (indicated by values in bold). Notably, UniChart with GCL shows a significant performance improvement, achieving up to a 9.04\% increase on ChartQA and a 7.59 increase on OpenCQA compared to the results from the provided Hugging Face checkpoint. Additionally, UniChart with GCL surpasses the original paper's results by 1.2\% on ChartQA and 1.47 on OpenCQA. While VL-T5 with GCL does not achieve the highest performance on ChartQA, this does not imply that GCL diminishes its performance because the graph integration method used differs from that in \cite{dai2024msg}. Overall, both UniChart and VL-T5 attain their best scores on both ChartQA and OpenCQA when incorporating both graph and graph contrastive learning, highlighting the effectiveness of our proposed approach.

The performance of our proposed CoT prompting method on MLLMs is summarized in Table~\ref{tab:cot_combined}. The term ``Original" refers to using the questions directly from the dataset, while ``Prompt 1-3" denotes the prompts outlined in Table~\ref{tab:cqaprompt}. In ChartQA, we only utilize the augmentation set, as MLLMs have acquired more mathematical reasoning ability during pre-training, resulting in higher performance on the human set. This, however, is not a fair comparison to fine-tuned models with different backbones. As shown, MLLMs perform better on ChartQA than ChartX across various prompt settings. A possible explanation is the difference in chart type diversity: ChartQA includes only three chart types, while ChartX features 18, making the latter more complex and highlighting the limitations of MLLMs in handling intricate chart types.

% Our proposed CoT prompting method shows greater promise on ChartQA. In ChartQA, the performance of both Qwen2-VL and Llama improves with the CoT prompt. In contrast, for ChartX, only Qwen2-VL shows improved performance. A possible explanation for this discrepancy is the difference in chart type diversity: ChartQA includes only 3 chart types, whereas ChartX features 18, making the latter more challenging.

% As seen in Table~\ref{tab:gcl_all}, in contrast to the 9.04\% improvement observed with fine-tuned UniChart, CoT prompting on MLLMs results in only a modest increase in average precision for ChartQA, with an improvement of just 1.72\%. This suggests that our proposed scene graph, coupled with the GCL-based prompt, introduces more informative knowledge to the backbone model.

As CoT prompting does not constantly improve performance, we use Qwen2-VL as a baseline for comparison in Table~\ref{tab:gcl_all}, as it consistently achieves relatively high performance across all prompt settings. Another observation is the low BLEU score in OpenCQA for Qwen2-VL, with a maximum of only 0.77. This is because of the significant difference between the generated answers and the ground truth regarding grammar and writing style. This finding highlights the limitations of using MLLMs for explanation tasks and underscores the necessity of designing more suitable evaluation metrics.

% \begin{table}[h]
% \centering
% \small
% \begin{tabular}{@{}p{0.10\linewidth}%
%                 p{0.20\linewidth}%
%                 >{\centering\arraybackslash}p{0.10\linewidth}%
%                 >{\centering\arraybackslash}p{0.12\linewidth}%
%                 >{\centering\arraybackslash}p{0.12\linewidth}%
%                 >{\centering\arraybackslash}p{0.12\linewidth}@{}}
% \toprule
% \textbf{Dataset} & \textbf{Models} & \textbf{Original} & \textbf{Prompt 1} & \textbf{Prompt 2} & \textbf{Prompt 3} \\ \midrule
% \multirow{3}{*}{\textbf{ChartQA}}  
%     & Qwen2-VL(7B)      & \textbf{76.28}  & \textbf{78.00} & \textbf{77.44}  & \textbf{77.16}  \\
%     & ChartLlama(13B)   & 71.92 & 66.24  & 65.72  & 63.60  \\
%     & Llama 3(11B)      & 72.88  & 75.60  & 74.92  & 76.12  \\ \midrule
% \multirow{3}{*}{\textbf{ChartX}}   
%     & Qwen2-VL(7B)      & 30.21  & 33.77 & 31.68  & 31.25  \\
%     & ChartLlama(13B)   & 16.32 & 15.54  & 14.32  & 13.98  \\
%     & Llama 3(11B)      & \textbf{44.36} & \textbf{39.50}  & \textbf{39.24}  & \textbf{40.36}  \\ 
% \bottomrule
% \end{tabular}
% \caption{Chain of Thought (CoT) Performance on ChartQA and ChartX}
% \label{tab:cot_combined}
% \vspace{-0.5cm}
% \end{table}

\begin{table}[h]
\centering
\small
\begin{tabular}{@{}p{0.10\linewidth}%
                p{0.20\linewidth}%
                >{\centering\arraybackslash}p{0.10\linewidth}%
                >{\centering\arraybackslash}p{0.12\linewidth}%
                >{\centering\arraybackslash}p{0.12\linewidth}%
                >{\centering\arraybackslash}p{0.12\linewidth}@{}}
\toprule
\textbf{Dataset} & \textbf{Models} & \textbf{Original} & \textbf{Prompt 1} & \textbf{Prompt 2} & \textbf{Prompt 3} \\ \midrule
\multirow{3}{*}{\textbf{\shortstack{ChartQA \\ aug. set}}}  
    & Qwen2-VL(7B)      & 85.60  & \textbf{84.56} & \textbf{84.16}  & \textbf{84.64}  \\
    & ChartLlama(13B)   & \textbf{90.96} & 83.20  & 82.40  & 79.68  \\
    & Llama 3(11B)  & 72.64  & 77.12  & 76.64  & 76.00  \\ \midrule
\multirow{3}{*}{\textbf{ChartX}}   
    & Qwen2-VL(7B)      & 30.21  & 33.77 & 31.68  & 31.25  \\
    & ChartLlama(13B)   & 16.32 & 15.54  & 14.32  & 13.98  \\
    & Llama 3(11B)      & \textbf{44.36} & \textbf{39.50}  & \textbf{39.24}  & \textbf{40.36}  \\ 
\bottomrule
\end{tabular}
\caption{Chain of Thought (CoT) Performance on ChartQA and ChartX}
\label{tab:cot_combined}
\vspace{-0.5cm}
\end{table}

% \begin{table}[h]
% \centering
% \small
% \begin{tabular}{@{}p{0.10\linewidth}%
%                 p{0.20\linewidth}%
%                 >{\centering\arraybackslash}p{0.10\linewidth}%
%                 >{\centering\arraybackslash}p{0.12\linewidth}%
%                 >{\centering\arraybackslash}p{0.12\linewidth}%
%                 >{\centering\arraybackslash}p{0.12\linewidth}@{}}
% \toprule
% \textbf{Dataset} & \textbf{Models} & \textbf{Original} & \textbf{Prompt 1} & \textbf{Prompt 2} & \textbf{Prompt 3} \\ \midrule
% \multirow{3}{*}{\textbf{ChartQA}} & Qwen2-VL(7B)    & 
% \textbf{76.28}          & \textbf{78.00}         & \textbf{77.44}           & \textbf{77.16}       \\
%                          & ChartLlama(13B)  & 71.92          & 66.24           & 65.72       & 63.60            \\
%                          & Llama 3(11B)     & 72.88          & 75.60            & 74.92           & 76.12           \\ \midrule
% \multirow{3}{*}{\textbf{ChartX}}  & Qwen2-VL(7B)    & \textbf{30.21}          & \textbf{33.77}       & \textbf{31.68}       & \textbf{31.25}       \\
%                          & ChartLlama(13B)  & 16.32          & 15.54           & 14.32           & 13.98           \\
%                          & Llama 3(11B)     & 44.36          & 39.50           & 39.24           & 40.36           \\ \bottomrule
% \end{tabular}
% \caption{Chain of Thought (CoT) Performance on ChartQA and ChartX}
% \label{tab:cot_performance}
% \end{table}

\subsection{Ablation Study}
In addition to applying GCL across multimodal graphs, we also experimented with GCL within a single graph, as described in \cite{DBLP:conf/iclr/VelickovicFHLBH19,DBLP:conf/www/0001XYLWW21}. Specifically, for the graphs $G_v$ and $G_t$, we randomly removed $p = 30\%$ of the edges to generate new graphs $\widetilde{G_v}$ and $\widetilde{G_t}$. Using $GCN_v$ and $GCN_t$, we then extracted the augmented representations $\widetilde{H_v} = GCN_v(\widetilde{G_v})$ and $\widetilde{H_t} = GCN_t(\widetilde{G_t})$. For GCL within a single-modality graph, the positive node pairs were defined as nodes representing the same object in both the original and augmented graphs. This approach allowed us to extend the loss function from Equation \ref{eq:loss} as follows:
\begin{equation}
    \begin{split}
    \ell_{cl} & = \ell_v(u_i, v_i \mid u \in H_v, v \in \widetilde{H_v}) \\ 
    & + \ell_t(u_i, v_i \mid u \in H_t, v \in \widetilde{H_t})  + \ell_{vt}(u_i, v_i \mid u \in H_v, v \in H_t)
    \end{split}
\end{equation}

Here, $\ell_v$ and $\ell_t$ correspond to the contrastive loss within single-modality graphs $G_v$ and $G_t$, while $\ell_{vt}$ captures the contrastive loss between the two modalities. We present the results with and without intra-modality GCL in Table \ref{tab:intra-cl}. From the table, we observe that intra-modality GCL decreases the performance of both UniChart and VL-T5 on ChartQA and OpenCQA, indicating that inter-modality GCL alone is sufficient for multimodal graph learning in generation tasks.

\begin{table}[h]
\centering
\small
\begin{tabular}{@{}p{0.10\linewidth}>{\centering\arraybackslash}p{0.10\linewidth}>{\centering\arraybackslash}p{0.10\linewidth}>{\centering\arraybackslash}p{0.10\linewidth}>{\centering\arraybackslash}p{0.10\linewidth}>{\centering\arraybackslash}p{0.10\linewidth}>{\centering\arraybackslash}p{0.10\linewidth}@{}}
\toprule
\multirow{2}{*}{\textbf{Model}} & \multirow{2}{*}{\shortstack{\textbf{INTRA-}\\\textbf{CL}}} & \multirow{2}{*}{\shortstack{\textbf{INTER-}\\\textbf{CL}}} & \multicolumn{3}{c}{\textbf{ChartQA}} & \multicolumn{1}{c}{\textbf{OpenCQA}} \\ 
& & & \textbf{aug.} & \textbf{human} & \textbf{avg.} & \textbf{BLEU} \\ \midrule
\multirow{2}{*}{UniChart} & $\times$ & $\checkmark$ & \textbf{90.00} & \textbf{44.88} & \textbf{67.44} & \textbf{16.35} \\
& $\checkmark$ & $\checkmark$ & 89.12 & 41.52 & 65.32 & 15.89 \\ \midrule
\multirow{2}{*}{VL-T5} & $\times$ & $\checkmark$ & \textbf{92.00} & \textbf{34.96} & \textbf{63.6} & \textbf{18.42} \\
& $\checkmark$ & $\checkmark$ & 90.08 & 34.08 & 62.08 & 18.38 \\ \bottomrule
\end{tabular}
\caption{Performance Comparison of Models With and Without Intra-CL (CL: Contrastive Learning)}
\label{tab:intra-cl}
\vspace{-0.5cm}
\end{table}

\begin{table}[h]
\centering
\small
\renewcommand{\arraystretch}{1.2} % Adjust line spacing
\begin{tabular}{@{}p{0.10\linewidth}>{\centering\arraybackslash}p{0.10\linewidth}>{\centering\arraybackslash}p{0.10\linewidth}>{\centering\arraybackslash}p{0.10\linewidth}>{\centering\arraybackslash}p{0.10\linewidth}>{\centering\arraybackslash}p{0.10\linewidth}>{\centering\arraybackslash}p{0.10\linewidth}@{}}
\toprule
\multirow{2}{*}{\textbf{Model}} & \multirow{2}{*}{\textbf{VG}} & \multirow{2}{*}{\textbf{TG}} & \multicolumn{3}{c}{\textbf{ChartQA}} & \multicolumn{1}{c}{\textbf{OpenCQA}} \\
& & & \textbf{aug.} & \textbf{human} & \textbf{avg.} & \textbf{BLEU} \\ \midrule
\multirow{3}{*}{Unichart} & $\checkmark$ & $\checkmark$ & \textbf{90.00} & \textbf{44.88} & \textbf{67.44} & 16.35 \\
 & $\times$ & $\checkmark$ & 89.20 & 44.08 & 66.64 & 16.38 \\
 & $\checkmark$ & $\times$ & 88.96 & 44.72 & 66.84 & \textbf{16.68} \\
 \midrule
\multirow{3}{*}{VL-T5} & $\checkmark$ & $\checkmark$ & \textbf{92.00} & 34.96 & 63.60 & 18.42 \\
 & $\times$ & $\checkmark$ & 91.52 & 37.68 & 64.60 & 18.65 \\
 & $\checkmark$ & $\times$ & 91.92 & \textbf{38.16} & \textbf{65.04} & \textbf{19.03} \\
 \bottomrule
\end{tabular}%
\caption{Performance Comparison with Respect to Different Graph Settings (VG: Visual Graph, TG: Textual Graph)}
\label{tab:vgtgp}
\vspace{-0.5cm}
\end{table}

Afterwards, we aimed to understand the importance of both graphs in our multimodal GCL. Instead of utilizing both visual and textual graphs, we feed either the textual or visual graph as the prompt to the decoder and presented the results in Table \ref{tab:vgtgp}. The table shows that using visual and textual graphs together does not always yield the best performance. In most cases, models using only the visual graph as a soft prompt achieved the best results. One possible reason is that the predicted labels from Mask-RCNN are highly noisy, particularly on OpenCQA, as it is not trained on the dataset. However, given that UniChart with multimodal GCL achieved the highest performance on ChartQA, we believe our proposed method has significant potential for further improvement. A more sophisticated GCL method design will be explored in future work. 
\section{Conclusion}
Chart question answering is a key task in chart understanding, with automated models enhancing productivity. However, transformer-based models often lose object-level information due to the fragmentation of images into patches. We address this by integrating a multimodal graph layer and adopting a contrastive learning framework, which improves performance on both ChartQA and OpenCQA datasets. Additionally, while recent pre-trained multimodal large language models (MLLMs) show promise, they face hallucination issues. We explore chain-of-thought prompting to mitigate this, though further investigation is needed for more significant gains.

\bibliographystyle{ACM-Reference-Format}
\bibliography{sample-base}

\end{document}